\documentclass{article}

\usepackage[final]{nips_2016}

\usepackage[utf8]{inputenc} % allow utf-8 input
\usepackage[T1]{fontenc}    % use 8-bit T1 fonts
\usepackage{hyperref}       % hyperlinks
\usepackage{url}            % simple URL typesetting
\usepackage{booktabs}       % professional-quality tables
\usepackage{amsfonts}       % blackboard math symbols
\usepackage{nicefrac}       % compact symbols for 1/2, etc.
\usepackage{microtype}      % microtypography

\usepackage{framed}
\usepackage[pdftex]{graphicx}
\usepackage[usenames, dvipsnames]{color}
\usepackage{subcaption}
\usepackage{tabularx}

\usepackage{slashbox}
\usepackage{colortbl}
\usepackage{array}

\title{Generative Deep Neural Networks for Dialogue: \\
A Short Review}

\author{
  Iulian Vlad Serban \\ Department of Computer Science \\ and Operations Research, \\ University of Montreal \\ 
  \And
  Ryan Lowe \\ School of Computer Science, \\ McGill University
  \And
  Laurent Charlin  \\ School of Computer Science, \\ McGill University
  \And  
  Joelle Pineau \\  School of Computer Science, \\ McGill University
}

\begin{document}

\maketitle

\begin{abstract}
Researchers have recently started investigating deep neural networks for dialogue applications.
In particular, generative sequence-to-sequence (Seq2Seq) models have shown promising results for unstructured tasks, such as word-level dialogue response generation.
The hope is that such models will be able to leverage massive amounts of data to learn meaningful natural language representations and response generation strategies, while requiring a minimum amount of domain knowledge and hand-crafting.
An important challenge is to develop models that can effectively incorporate dialogue context and generate meaningful and diverse responses.
In support of this goal, we review recently proposed models based on generative encoder-decoder neural network architectures, and show that these models have better ability to incorporate long-term dialogue history, to model uncertainty and ambiguity in dialogue, and to generate responses with high-level compositional structure.
\end{abstract}

\vspace*{-0.5\baselineskip}
\section{Introduction}
\vspace*{-0.5\baselineskip}

Researchers have recently started investigating sequence-to-sequence (Seq2Seq) models for dialogue applications.
These models typically use neural networks to both represent dialogue histories and to generate or select appropriate responses. %, which are trained to mimic dialogues occurring in a dialogue corpus.
Such models are able to leverage large amounts of data in order to learn meaningful natural language representations and generation strategies, while requiring a minimum amount of domain knowledge and hand-crafting.
Although the Seq2Seq framework is different from the well-established goal-oriented setting~\citep{gorin1997may,young2000probabilistic,singh2002optimizing},
these models have already been applied to several real-world applications, with Microsoft's system Xiaoice~\citep{markoff2015forsymp} and Google's Smart Reply system~\citep{kannan2016smart} as two prominent examples.

Researchers have mainly explored two types of Seq2Seq models.
The first are generative models,
which are usually trained with cross-entropy to generate responses word-by-word conditioned on a dialogue context~\citep{ritter2011data,vinyals2015neural,sordoni2015aneural,shang2015neural,li2015diversity,DBLP:conf/aaai/SerbanSBCP16}.
The second are discriminative models, which are trained to select an appropriate response from a set of candidate responses \citep{lowe2015ubuntu,bordes2016learning,inaba2016neural,yu2016strategy}.
In a related strand of work, researchers have also investigated applying neural networks to the different components of a standard dialogue system, including natural language understanding, natural language generation, dialogue state tracking and evaluation \citep{wen2016network,wen-EtAl:2015:EMNLP,henderson2013deep,mrkvsic2015multi,su2015learning}. %wen2015stochastic
In this paper, we focus on generative models trained with cross-entropy.
%Our work focuses on generative model trained with the cross-entropy training criterion.

One weakness of current generative models is their limited ability to incorporate rich dialogue context and to generate meaningful and diverse responses \citep{DBLP:conf/aaai/SerbanSBCP16,li2015diversity}.
To overcome this challenge, we propose new generative models that are better able to incorporate long-term dialogue history, to model uncertainty and ambiguity in dialogue, and to generate responses with high-level compositional structure.
Our experiments demonstrate the importance of the model architecture and the related inductive biases in achieving this improved performance.

% focus on the generative models, because they can easily be scaled to large datasets and open domains.

\begin{figure}[ht]
  \centering
  \includegraphics[width=0.8\textwidth]{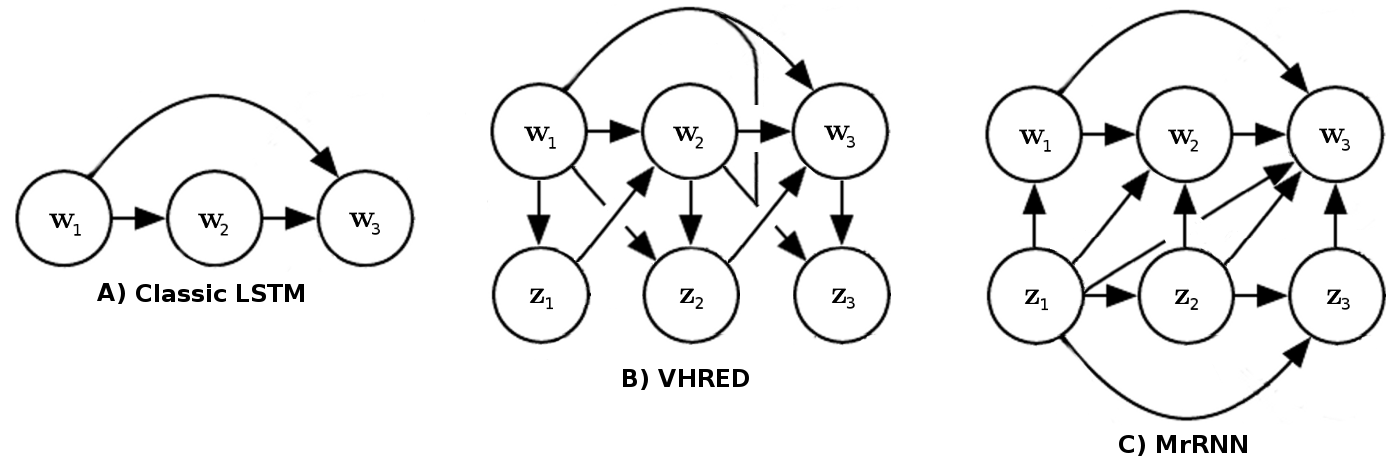}
  \caption{Probabilistic graphical models for dialogue response generation.
  Variables $\mathbf{w}$ represent natural language utterances.
  Variables $\mathbf{z}$ represent discrete or continuous stochastic latent variables.
  (A): Classic LSTM model, which uses a shallow generation process. This is problematic because it has no mechanism for incorporating uncertainty and ambiguity and because it forces the model to generate compositional and long-term structure incrementally on a word-by-word basis.
  (B): VHRED expands the generation process by adding one latent variable for each utterance, which helps incorporate uncertainty and ambiguity in the representations and generate meaningful, diverse responses.
  (C): MrRNN expands the generation process by adding a sequence of discrete stochastic variables for each utterance, which helps generate responses with high-level compositional structure.}
  \label{fig:MultiresolutionHRED}
\end{figure}

\vspace*{-0.5\baselineskip}
\section{Models}
\vspace*{-0.5\baselineskip}

% , each with its own inductive biases motivated by linguistic observations:
\textbf{HRED}: The Hierarchical Recurrent Encoder-Decoder model (HRED) \citep{DBLP:conf/aaai/SerbanSBCP16} is a type of Seq2Seq model that decomposes a dialogue into a two-level hierarchy: a sequence of utterances, each of which is a sequence of words.
HRED consists of three recurrent neural networks (RNNs): an \textit{encoder} RNN, a \textit{context} RNN and a \textit{decoder} RNN.
Each utterance is encoded into a real-valued vector representation by the \textit{encoder} RNN.
These utterance representations are given as input to the \textit{context} RNN, which computes a real-valued vector representation summarizing the dialogue at every turn.
This summary is given as input to the \textit{decoder} RNN, which generates a response word-by-word.
Unlike the RNN encoders in previous Seq2Seq models, the \textit{context} RNN is only updated once every dialogue turn and uses the same parameters for each update.
This gives HRED an inductive bias that helps incorporate long-term context and learn invariant representations.

\textbf{VHRED}: The Latent Variable Hierarchical Recurrent Encoder-Decoder model (VHRED) \citep{serban2016hierarchical} is an HRED model with an additional component: a high-dimensional stochastic latent variable at every dialogue turn.
As in HRED, the dialogue context is encoded into a vector representation using \textit{encoder} and \textit{context} RNNs.
%each utterance is encoded into a vector representation by an \textit{encoder} RNN.
%These utterance representations are given as input to the \textit{context} RNN, which computes a vector representation summarizing the dialogue at every turn.
Conditioned on the summary vector at each dialogue turn, VHRED samples a multivariate Gaussian variable, which is given along with the summary vector as input to the \textit{decoder} RNN.
%which generates the response word-by-word.
The multivariate Gaussian latent variable allows modelling ambiguity and uncertainty in the dialogue through the latent variable distribution parameters (mean and variance parameters).
This provides a useful inductive bias, which helps VHRED encode the dialogue context into a real-valued embedding space even when the dialogue context is ambiguous or uncertain, and it helps VHRED generate more diverse responses.

\textbf{MrRNN}: The Multiresolution RNN (MrRNN) \citep{serban2016multiresolution} models dialogue as two parallel stochastic sequences:
a sequence of high-level coarse tokens (coarse sequences), and a sequence of low-level natural language words (utterances).
The coarse sequences follow a latent stochastic process---analogous to hidden Markov models---which conditions the utterances through a hierarchical generation process.
The hierarchical generation process first generates the coarse sequence, and conditioned on this generates the natural language utterance.
In our experiments, the coarse sequences are defined as either noun sequences or activity-entity pairs (predicate-argument pairs) extracted from the natural language utterances.
The coarse sequences and utterances are modelled by two separate HRED models.
The hierarchical generation provides an important inductive bias, because it helps MrRNN model high-level, compositional structure and generate meaningful and on-topic responses.

%We implement all models in Theano.
%~\cite{2016arXiv160502688short}.
%We train all models w.r.t.\@ the log-likelihood or joint log-likelihood on the training set using Adam~\cite{kingma2014adampublished}. 
%The models are trained using early stopping with patience based on the validation set log-likelihood.
%W%e choose model hyperparameters -- such as the number of hidden units, word embedding dimensionality, and learning rate -- based on the validation set log-likelihood.
%For further details on the models, see %\cite{serban2016multiresolution,serban2016hierarchical,DBLP:journals/corr/SerbanLCP15}.

\vspace*{-0.5\baselineskip}
\section{Experiments}
\vspace*{-0.5\baselineskip}

We apply our generative models to dialogue response generation on the Ubuntu Dialogue Corpus~\citep{lowe2015ubuntu}.
For each example, given a dialogue context, the model must generate an appropriate response.
We also present results on Twitter in the Appendix.
This task has been studied extensively in the recent literature \citep{ritter2011data,sordoni2015aneural,li2015diversity}.
% DBLP:conf/aaai/SerbanSBCP16
% lowe2015ubuntu

\newcolumntype{L}[1]{>{\raggedright\arraybackslash}p{#1}}

\begin{table*}[t]
 \caption{Model response examples. The arrows indicate change of turn.}
 \label{table:ubuntu-examples-small}
 \scriptsize
 \centering
 \begin{tabular}{L{80mm}L{50mm} }
 \textbf{Dialogue Context} & \textbf{Model Response}  \\ \hline
        Hello I tried to install ubuntu studio but I get W : Failed to fetch \textless url \textgreater \ 404 Not Found when apt-get update $\rightarrow$ the ppa you added is not online & \textbf{MrRNN Act.\@ -Ent.\@:} try a different mirror \newline \textbf{MrRNN Noun:} something \newline \textbf{VHRED:} but I dont know how to add it \newline \textbf{HRED:} how do I do that ? \\ \hline 
        hi community.. is there any difference between " ~/.bashrc " " ~/.bashrc. local " $\rightarrow$ yes, they're different files :-) the default ~/.bashrc sources ~/.bashrc. local $\rightarrow$ sorry I could undersatnd... If I write on terminal " gedit ~/.bashrc " it opens a file but how I open " ~/.bashrc. local " $\rightarrow$ " gedit ~/.bashrc. local "... $\rightarrow$ but it just open blank file nothing inside it & \textbf{MrRNN Act.\@-Ent.\@:} open up gedit ~/.bashrc, there is a file called " something " \newline \textbf{MrRNN Noun:} the file is empty.. \newline \textbf{VHRED:} it's in ~/.bashrc or ~/.bashrc  \newline \textbf{HRED:} what are you trying to do ? \\ \hline
         Hey guys what do you in general use for irc something ubuntu xchat or xchat-gnome ? $\rightarrow$ without -gnome. that is just cut down $\rightarrow$ you mean drop xchat-gnome and go with xchat ? & \textbf{MrRNN Act.\@ -Ent.\@:} im using xchat right now \newline \textbf{MrRNN Noun:} what is xchat-gnome ? \newline \textbf{VHRED:} correct \quad \textbf{HRED:} yes \\ \hline
        when setting up rules with iptables command only writes changes this file " \/etc\/iptables. rules "? i ask this so i can backup before messing anything $\rightarrow$ sudo iptables-save something . dat to backup your rules restore with sudo iptables-restore \textless \ something . dat & \textbf{MrRNN Act.\@ -Ent.\@:} I don't see any reason why you need iptables to do that \newline \textbf{MrRNN Noun:} are you using ubuntu ? \newline \textbf{VHRED:} thx \quad \textbf{HRED:} thanks \\ 
 \bottomrule
 \end{tabular}
\end{table*}

\textbf{Corpus}: 
% developed by Lowe et al.
%\footnote{We use the Ubuntu Dialogue Corpus v2.0 extracted Jamuary, 2016: \url{http://cs.mcgill.ca/~jpineau/datasets/ubuntu-corpus-1.0/}}\footnote{For the context and response pairs in the official Ubuntu Dialogue Corpus, there is no distinction between the users having technical problems and other users helping them resolve their problems. Therefore the model must learn to act as both a user in need of technical support and as a user providing technical support.}
The Ubuntu Dialogue Corpus consists of about half a million dialogues extracted from the \textit{\#Ubuntu} Internet Relayed Chat (IRC) channel.
Users entering this chat channel usually have a specific technical problem. 
Typically, users first describe their problem, and other users try to help them resolve it.
The technical problems range from software-related and hardware-related issues (e.g.\@ installing packages, fixing broken drivers) to informational needs (e.g.\@ finding software).

\textbf{Evaluation}: 
%Accurate evaluation of dialogue system responses is difficult~\cite{schatzmann2005quantitative}.
%Liu et al.~\cite{liu2016not} have recently shown that all automatic evaluation metrics adopted for such evaluation, including word overlap-based metrics such as BLEU and METEOR, have either very low or no correlation with human judgement of system performance.
We carry out an in-lab human study to evaluate the model responses.
We recruit 5 human evaluators. We show each evaluator between 30 and 40 dialogue contexts with the ground truth response, and 4 candidate model responses.
% (HRED, HRED + Activity-Entity Features, MrRNN Noun, and MrRNN Activity-Entity).
For each example, we ask the evaluators to compare the candidate responses to the ground truth response and dialogue context,
and rate them for fluency and relevancy on a scale 0--4, where 0 means incomprehensible or no relevancy and 4 means flawless English or all relevant.
In addition to the human evaluation, we also evaluate dialogue responses w.r.t.\@ the activity-entity metrics proposed by ~\cite{serban2016multiresolution}. These metrics measure whether the model response contains the same activities (e.g.\@ download, install) and entities (e.g.\@ ubuntu, firefox) as the ground truth responses.
Models that generate responses with the same activities and entities as the ground truth responses---including expert responses, which often lead to solving the user's problem---are given higher scores.  Sample responses from each model are shown in Table~\ref{table:ubuntu-examples-small}.

\begin{table}[ht]
  \caption{Ubuntu evaluation using F1 metrics w.r.t.\@ activities and entities (mean scores $\pm \ 90\%$ confidence intervals), and human fluency and human relevancy scores given on a scale 0-4 {\small ($^*$ indicates scores significantly different from baseline models at $90\%$ confidence)}} \label{tabel:ubuntu_results}
  \small
  \centering
    \begin{tabular}{lccccccc}
    \toprule
%     & \textbf{Activity} & \textbf{Entity} & \multicolumn{2}{c}{\textbf{Human Eval.\@}} \\ \midrule
    \textbf{Model} & \textbf{F1 Activity} & \textbf{F1 Entity} & \textbf{Human Fluency} & \textbf{Human Relevancy} \\
    \midrule
        LSTM & \small $1.18$ \footnotesize $\pm 0.18$ & \small $0.87$ \footnotesize $\pm 0.15$ & - & - \\
        HRED & \small $4.34$ \footnotesize $\pm 0.34$ & \small $2.22$ \footnotesize $\pm 0.25$ & 2.98 & 1.01\\
        VHRED & \small $4.63$ \footnotesize $\pm 0.34$ & \small $2.53$ \footnotesize $\pm 0.26$ & - & - \\       
        %\parbox[c][2.65em][c]{0.070\textwidth}{HRED + \\ Act.\@-Ent.\@} & \small $5.46$ \footnotesize $\pm 0.35$ & \small $2.44$ \footnotesize $\pm 0.26$ & 2.96 & 0.75 \\
%        \parbox[c][2.5em][c]{0.11\textwidth}{MrRNN \\ Stem.\@} & \small $1.66$ & \small $3.18$ & - & - \\
       MrRNN Noun & \small $4.04$ \footnotesize $\pm 0.33$ & \small $\mathbf{6.31}$ \footnotesize $\mathbf{\pm 0.42}$ & \small $\mathbf{3.48}^*$ & \small $\mathbf{1.32}^*$ \\
        MrRNN Act.\@-Ent.\@ & \small $\mathbf{11.43}$ \footnotesize $\pm \mathbf{0.54}$ & \small $3.72$ \footnotesize $\pm 0.33$ & \small $\mathbf{3.42}^*$ & \small $1.04$ \\ \bottomrule
    \end{tabular}
\end{table}
% \parbox[c][2.65em][c]{0.070\textwidth}{VHRED}

\textbf{Results}:
The results are given in Table~\ref{tabel:ubuntu_results}.
The MrRNNs perform substantially better than the other models w.r.t.\@ both the human evaluation study and the evaluation metrics based on activities and entities.
MrRNN with noun representations obtains an F1 entity score at $6.31$, while all other models obtain less than half F1 scores between $0.87 - 2.53$, and human evaluators consistently rate its fluency and relevancy significantly higher than all the baseline models.
MrRNN with activity representations obtains an F1 activity score at $11.43$, while all other models obtain less than half F1 activity scores between $1.18-4.63$, and performs substantially better than the baseline models w.r.t. the\@ F1 entity score.
This indicates that the MrRNNs have learned to model high-level, goal-oriented sequential structure in the Ubuntu domain.
Followed by these, VHRED performs better than the HRED and LSTM models w.r.t.\@ both activities and entities.
This shows that VHRED generates more appropriate responses, which suggests that the latent variables are useful for modeling uncertainty and ambiguity.
Finally, HRED performs better than the LSTM baseline w.r.t.\@ both activities and entities,
which underlines the importance of representing longer-term context.
These conclusions are confirmed by additional experiments on response generation for the Twitter domain (see Appendix).
%, as well as manual inspections of the responses.

%the results indicate that the MrRNNs have learned to model high-level, goal-oriented sequential structure in the Ubuntu domain.
%In addition to this, the results indicate that modelling uncertainty and ambiguity through a latent variable has also helped the VHRED model generate appropriate responses.

%Furthermore, the MrRNNs outperform the HRED baseline augmented activity-entity features across all metrics.
%This indicates that the hierarchical generation process helps MrRNNs generate responses by adding high-level, compositional structure.

%Followed by the MrRNN models, the VHRED model performs better than both the HRED and LSTM models w.r.t.\@ both activities and entities.
%This indicates that the latent variable in VHRED helps generate meaningful responses.

% induces an inductive bias,

\vspace*{-0.5\baselineskip}
\section{Discussion}
\vspace*{-0.5\baselineskip}

% added a quick discussion, feel free to expand, remove or re-write

We have presented generative models for dialogue response generation.
%that are variants on the traditional Seq2Seq framework.
We have proposed architectural modifications with inductive biases towards 1) incorporating longer-term context, 2) handling uncertainty and ambiguity, and 3) generating diverse and on-topic responses with high-level compositional structure.
Our experiments show the advantage of the architectural modifications quantitatively through human experiments and qualitatively through manual inspections.
These experiments demonstrate the need for further research into generative model architectures.
Although we have focused on three generative models, other model architectures such as memory-based models \citep{bordes2016learning,weston2014memory} and attention-based models \citep{shang2015neural} have also demonstrated promising results and therefore deserve the attention of future research.

In another line of work, researchers have started proposing alternative training and response selection criteria~\citep{weston2016dialog}. \cite{li2015diversity} propose ranking candidate responses according to a mutual information criterion, in order to incorporate dialogue context efficiently and retrieve on-topic responses.
\cite{li2016deep} further propose a model trained using reinforcement learning to optimize a hand-crafted reward function. Both these models are motivated by the lack of \textit{diversity} observed in the generative model responses.
Similarly, \cite{yu2016strategy} propose a hybrid model---combining retrieval models, neural networks and hand-crafted rules---trained using reinforcement learning to optimize a hand-crafted reward function.
In contrast to these approaches, without combining several models or having to modify the training or response selection criterion, VHRED generates more diverse responses than previous models.
Similarly, by optimizing the joint log-likelihood over sequences, MrRNNs generate more appropriate and on-topic responses with compositional structure.
Thus, improving generative model architectures has the potential to compensate --- or even remove the need --- for hand-crafted reward functions. 

%override counter-act the need for 
%The need for further research into developing generative model architectures is 
%This underlines our previous conclusion, that much research is still required in developing generative model architectures.

At the same time, the models we propose are not necessarily better language models, which are more efficient at compressing dialogue data as measured by word perplexity.
Although these models produce responses that are preferred by humans, they often result in higher test set perplexity than traditional LSTM language models.
This suggests maximizing log-likelihood (i.e.\@ minimizing perplexity) is not a sufficient training objective for these models.
An important line of future work therefore lies in improving the objective functions for training and response selection, as well as learning directly from interactions with real users.

%may require external rewards from interacting with real users.
%This discrepancy between perplexity and human judgement of model responses 

%This discrepancy is because perplexity is reduced most significantly by the efficient compression of frequent (generic) utterances; since our models have inductive biases towards producing more diverse responses, they place lower probabilities on generic utterances, which results in a higher perplexity. 
%This suggests that maximizing log-likelihood (i.e.\@ minimizing perplexity) is not a sufficient training objective for these models, and we may require external rewards from interacting with real users.
%% insert paragraph about important future directions?

\newpage
\bibliographystyle{abbrvnat}
\begingroup
    \vspace{-2.0mm}
%    \scriptsize
    \small
    \setlength{\bibsep}{2pt}
    \bibliography{references}
\endgroup

\newpage
\section*{Appendix}

\subsection*{Twitter Results}

\textbf{Corpus}: We experiment on a Twitter Dialogue Corpus~\citep{ritter2011data} containing about one million dialogues.
The task is to generate utterances to append to existing Twitter conversations.
This task is typically categorized as a non-goal-driven task, because any fluent and on-topic response may be adequate.
%The dataset is extracted using a procedure similar to Ritter et al.~\shortcite{ritter2011data}, and is split into training, validation and test sets, containing respectively 749,060, 93,633 and 10,000 dialogues each.

\textbf{Evaluation}: We carry out a human study on Amazon Mechanical Turk (AMT).
We show human evaluators a dialogue context along with two potential responses:
one response generated from each model conditioned on the dialogue context.
We ask evaluators to choose the response most appropriate to the dialogue context.
If the evaluators are indifferent, they can choose neither response.
For each pair of models we conduct two experiments: one where the example contexts contain at least $80$ unique tokens (\emph{long context}),
and one where they contain at least $20$ (not necessarily unique) tokens (\emph{short context}).
We experiment with the LSTM, HRED and VHRED models, as well as a TF-IDF retrieval-based baseline model.
We do not experiment with the MrRNN models, because we do not have appropriate coarse representations for this domain.

\textbf{Results}:
The results given in Table~\ref{table:human-study-twitter} show that VHRED is strongly preferred in the majority of the experiments.
In particular, VHRED is strongly preferred over the HRED and TF-IDF baseline models for both short and long context settings. VHRED is also strongly preferred over the LSTM baseline model for long contexts, although the LSTM model is preferred over VHRED for short contexts.For short contexts, the LSTM model is often preferred over VHRED because the LSTM model tends to generate very \textit{generic} responses. Such \textit{generic} or \textit{safe} responses are reasonable for a wide range of contexts, but are not useful when applied through-out a dialogue, because the user would loose interest in the conversation.
%; however, the LSTM is preferred over VHRED for short contexts. %\footnote{The difference in preferences for VHRED vs. LSTM with long contexts is significant with $95\%$ confidence.}

In conclusion, VHRED performs substantially better overall than competing models,
which suggests that the high-dimensional latent variables help model uncertainty and ambiguity in the dialogue context and help generate meaningful responses.

\begin{table}[!htb]
  \caption{Wins, losses and ties (in \%) of VHRED against baselines based on the human study (mean preferences $\pm \ 90\%$ confidence intervals, where $*$ indicates significant differences at $90\%$ confidence) }
  \label{table:human-study-twitter}
  \def\arraystretch{1}
  \small
  \centering
  \begin{tabular}{lccc}
    \toprule
    \textbf{Opponent} & \textbf{Wins} & \textbf{Losses} & \textbf{Ties} \\ \midrule
    \multicolumn{4}{l}{\textbf{Short Contexts}}  \\ \midrule
    VHRED vs LSTM & \small $32.3$ \footnotesize $\pm 2.4$ & \small $\mathbf{42.5}$ \footnotesize $\mathbf{\pm 2.6}^{*}$ & \small $25.2$ \footnotesize $\pm 2.3$ \\
    VHRED vs HRED & \small $\mathbf{42.0}$ \footnotesize $\mathbf{\pm 2.8}^{*}$ & \small $31.9$ \footnotesize $\pm 2.6$ & \small $26.2$ \footnotesize $\pm 2.5$ \\
    VHRED vs TF-IDF & \small $\mathbf{51.6}$ \footnotesize $\mathbf{\pm 3.3}^{*}$ & \small $17.9$ \footnotesize $\pm 2.5$ & \small $30.4$ \footnotesize $\pm 3.0$ \\ \midrule
    \multicolumn{4}{l}{\textbf{Long Contexts}}  \\ \midrule
    VHRED vs LSTM & \small $\mathbf{41.9}$ \footnotesize $\mathbf{\pm 2.2}^{*}$ & \small $36.8$ \footnotesize $\pm 2.2$ & \small $21.3$ \footnotesize $\pm 1.9$ \\
    VHRED vs HRED & \small $\mathbf{41.5}$ \footnotesize $\mathbf{\pm 2.8}^{*}$ & \small $29.4$ \footnotesize $\pm 2.6$ & \small $29.1$ \footnotesize $\pm 2.6$ \\
    VHRED vs TF-IDF & \small $\mathbf{47.9}$ \footnotesize $\mathbf{\pm 3.4}^{*}$ & \small $11.7$ \footnotesize $\pm 2.2$ & \small $40.3$ \footnotesize $\pm 3.4$ \\ \bottomrule
  \end{tabular}
 \end{table}

\end{document}